\pdfoutput=1

\documentclass[11pt]{article}

\usepackage[]{acl}

\usepackage{times}
\usepackage{latexsym}

\usepackage[T1]{fontenc}

\usepackage[utf8]{inputenc}

\usepackage{microtype}
\usepackage[compact]{titlesec}

\usepackage{multirow}
\usepackage{booktabs}
\usepackage{makecell}
\usepackage{graphicx}
\usepackage{enumitem}
\usepackage[normalem]{ulem}
\usepackage{adjustbox}
\usepackage[T1]{fontenc}
\usepackage[english]{babel}
\usepackage{microtype}
\usepackage{amsmath}
\usepackage{algorithm}
\usepackage{algpseudocode}
\usepackage{multirow}
\newtheorem{definition}{Definition}
\usepackage{enumitem}
\usepackage{comment}
\DeclareUnicodeCharacter{3B1}{\ensuremath{\alpha}}

\title{Zero-Shot Multi-Label Topic Inference with Sentence Encoders}

\author{Souvika Sarkar, Dongji Feng, Shubhra Kanti Karmaker Santu \\
Big Data Intelligence (BDI) Lab, Auburn University, Alabama, USA \\
\{szs0239, dzf0023, sks0086\}@auburn.edu}

\begin{document}
\maketitle
\begin{abstract}

Sentence encoders have indeed been shown to achieve superior performances for many downstream text-mining tasks and, thus, claimed to be fairly general. Inspired by this, we performed a detailed study on how to leverage these sentence encoders for the ``zero-shot topic inference'' task, where the topics are defined/provided by the users in \textit{real-time}. Extensive experiments on seven different datasets demonstrate that \textit{Sentence-BERT} demonstrates superior generality compared to other encoders, while \textit{Universal Sentence Encoder} can be preferred when efficiency is a top priority.

\end{abstract}

\section{Introduction}
As one of the most fundamental problems in text mining, topic modeling, and inference have been widely studied in the literature~\cite{alghamdi2015survey,jelodar2019latent,chauhan2021topic}. In this paper, we focus on \textit{Zero-shot} approaches~\cite{yin2019benchmarking,10.1145/2983323.2983866,veeranna2016using} for inferring topics from documents where document and topics were never  seen previously by a model. Furthermore, for developing \textit{Zero-shot} methods, we exclusively focus on leveraging the recent powerful sentence encoders.

The problem of zero-shot topic inference can be described using an intuitive example, where an end user (possibly a domain expert) is actively involved in the inference process. Consider that the domain expert is analyzing a large volume of health articles and wants to automatically infer topics of those articles with health-related topics like ``Autoimmune Disorders'', ``Heart health'', ``Arthritis'', etc. For this real-life use case, the user will provide the collection of documents as well as a set of topics to be used as topics for categorizing the documents. Additionally, the user may also provide a list of relevant keywords/clues associated with each topic which can be used as expert guidance for the inference process. The \textit{zero-shot topic inference} algorithm then infers topics for each document.

Naturally, zero-shot topic inference is a difficult task, and only limited previous works tackled this problem~\cite{yin2019benchmarking,10.1145/2983323.2983866,veeranna2016using}. However, recent developments in transfer learning research have demonstrated that pre-trained sentence embeddings like~\cite{cer2018universal,infersent,laser,sbert} can achieve promising results in many downstream zero-shot NLP tasks. Inspired by these, we focus on exploring zero-shot methods by using various sentence encoders for topic inference.

Thus, this paper aims to examine the transfer learning capabilities of popular sentence encoders [InferSent~\cite{infersent}, Language-Agnostic SEntence Representations (LASER)~\cite{laser}, Sentence-BERT (SBERT)~\cite{sbert}, and Universal Sentence Encoder (USE)~\cite{cer2018universal}] for topic inference tasks and subsequently, establish a benchmark for future study in this crucial direction. To achieve this, we conducted extensive experiments with multiple real-world datasets, including online product reviews, news articles, and health-related blog articles. We also implemented two topic-modeling and  word embeddings-based zero-shot approaches to compare as baselines.  Our experiment results show that among all four encoders \textit{Sentence-BERT} is superior in terms of generality compared to other encoders, while \textit{Universal Sentence Encoder} being the second best.

\section {Related Work} \label{section:RelatedWork}

This work is built upon prior research from multiple areas, including Topic Modeling and Categorization, ~\cite{blei2003latent,Wang2011StructuralTM,iwata2009modeling}, Text Annotation ~\cite{ogren2006knowtator, 10.1145/3331184.3331424, 10.1145/3397482.3450715},  Zero-Shot Learning ~\cite{veeranna2016using,yin2019benchmarking}, Sentence embeddings~\cite{casanueva2020efficient, DBLP:conf/emnlp/CerYKHLJCGYTSK18} etc. A brief discussion on each area and how this work is positioned concerning the state-of-the-art is as follows.

\subsection{Topic Modeling and Inference}

\smallskip

\textbf{Classical Unsupervised Topic Models}: Classical Topic Models emerged in the late '90s. Hofmann et. al.~\citet{hofmann1999probabilistic} proposed one of the early topic models, PLSA, which modeled word co-occurrence information under a probabilistic framework in order to discover the underlying semantic structure of the data. Later, ~\citet{blei2003latent} proposed Latent Dirichlet Allocation  (LDA), which extended PLSA by proposing a probabilistic model at the level of documents. To this date, LDA remains one of the most widely used topic models. Multiple works followed LDA later, including~\citet{Wang2011StructuralTM,du2013topic,he2016spatial,10.1145/2600428.2609565} etc.

\smallskip
\textbf{Topic Inference by Supervised Classification}: Several studies including~\cite{tuarob2015generalized,bundschus2009topic} have shown that it is possible to categorize topics from well-annotated collections of training data through supervised learning.~\citet{iwata2009modeling} proposed a topic model for analyzing and excerpting content-related categories from noisy annotated discrete data such as web pages stored in bookmarks. \citet{poursabzi2015speeding} combined document classification and topic models, where topic modeling was used to uncover the underlying semantic structure of documents in the collection.~\citet{engels2010automatic} came up with an automatic categorization scheme, in which they employed a latent topic model to generate topic distributions given a video and associated text. \citet{10.1145/3269206.3271737}, proposed a supervised method that addresses the lack of training data in neural text classification. Other researchers.~\citet{hassan2020towards} proposed a supervised classification problem for sexual violence report tracking.

\smallskip
\textbf{Zero-Shot Topic Inference}: Various topic modeling-based approaches have been explored for zero-shot topic inference; for instance, ~\citet{10.1145/3269206.3271671} worked towards a topic modeling approach for dataless text classification. 
Similarly, ~\citet{zha2019multi} proposed a novel Seed-guided Multi-label Topic Model based dataless text classification technique.~\citet{karmaker2016generative} proposed a zero-shot model that can mine implicit topics from online reviews without any supervised training. 

Researchers also explored the zero-shot topic inference paradigm using deep learning techniques where knowledge of topics is incorporated in the form of embeddings.~\citet{veeranna2016using}, adopted pre-trained word embedding for measuring semantic similarity between a label and documents. Further endeavor has been spent on zero-shot learning using semantic embedding by~\cite{hascoet2019semantic,zhang2019integrating,xie2021zero,rios2018few,yin2019benchmarking,xia2018zero,zhang2019integrating,pushp2017train,puri2019zero,yogatama2017generative,pushp2017train,chen2021zero,10.1145/3459637.3482403}.

\subsection{Sentence Embedding} 
Sentence encoders (Universal Sentence Encoders, InferSent, Language-Agnostic Sentence Representations, Sentence-BERT) are heavily in practice recently. In this section, we will discuss their applications in the research area.

The utility of these powerful sentence encoders has been tested for many popular NLP tasks, including Intent Classification~\citet{casanueva2020efficient}, Fake-News Detection~\citet{majumder2020detecting}, Duplicate Record Identification~\citet{lattar2020duplicate}, Humor detection~\citet{annamoradnejad2020colbert}, Ad-Hoc monitoring ~\citet{sarkar-covid} and COVID-19 Trending Topics Detection from tweets~\citet{asgaricovid}. Authors in ~\citet{10.1145/3404835.3463057} proposed a dual-view approach that enhances sentence embeddings. Another line of work focused on the performance of sentence embedding techniques for transfer-learning tasks~\cite{perone2018evaluation,enayet2020transfer}, whereas a group of researchers reported that state-of-the-art sentence embeddings are unable to capture sufficient information regarding sentence correctness and quality in the English language~\citet{rivas2019empirical, sarkar-etal-2022-exploring}. ~\citet{chen2019biosentvec} utilized USE to create domain-specific embeddings. Another school of researchers ~\cite{hassan2019bert, chen2018combining,tang2018improving} leveraged sentence embedding for recommending research articles and computing semantic similarity between articles. ~\citet{adi2016fine} proposed a framework that facilitated a better understanding of the encoded sentence representations and extended this work in ~\cite{adi2017analysis}, which discussed the effect of word frequency or word distance on the ability to encode sentences.

\subsection{Difference from Previous Works} 
Despite much research in this area, the latest sentence encoder's potential has not been systematically examined for the goal task. Although few previous works have leveraged sentence encoders for calculating the similarity between a text and a topic, most of the works so far have mainly focused on a single sentence encoder. In contrast, we study multiple state-of-the-art sentence encoders for the multilabel \textit{zero-shot} topic inference task and experiment with different ways of encoding both topics and documents, and thus, conduct a more comprehensive comparative study. We also propose a novel way to incorporate the auxiliary information provided by the user to encode topics, which eventually improved the inference result.

\section{\textbf{Problem Statement}}\label{sec:problem}

The traditional \textit{Topic Inference} task can be defined as follows: 

\begin{definition}
Given a collection of documents $D$ and a set of \textbf{pre-defined} topics $T$, infer one or more topics in $T$ FOR each document $d\in D$.
\end{definition} 

Thanks to the \textbf{pre-defined} set of topics $T$, the traditional \textit{Topic Inference} task can benefit from fine-tuning based on a carefully designed training set for supervised learning. On the other hand, we follow the idea of \textit{Definition-Wild Zero-Shot-Text Classification} coined by ~\citet{yin2019benchmarking}, which is as follows:

\begin{definition}
Definition-Wild 0SHOT-TC aims at learning a classifier f(·) : X → Y , where classifier f(·) never sees Y-specific labeled data in its model development.
\end{definition}

Extending on top of \textit{Definition-Wild (0SHOT-TC)}, we formalize our task from the user's standpoint in the following fashion:

\begin{definition}
Given a collection of documents $D=\{d_1, d_2,..., d_n\}$, a user $x$ and a set of \textbf{user-defined} topics $T_x=\{t_1, t_2,..., t_m\}$ provided in \textbf{real-time}, annotate each document $d_i\in D$ with zero or more topics from $T_x$ \textbf{without any further fine-tuning}.
\end{definition}

Note that, it is possible that two different users will provide a different set of topics for the same dataset based on their application needs and end goals. This essentially means creating customized training datasets beforehand is no longer possible because the target topics/labels are provided in real-time. We also assume that each topic $t$ is expressed as a word/phrase, and the user can provide a list of additional keywords $K_t$ associated with each topic $t$. In a nutshell, our ad-hoc problem setting assumes that the end user provides all the documents, the target topic, and an optional list of topic-related keywords as inputs in real time. The user here is usually a domain expert with specialized knowledge or skills in a particular area of endeavor (e.g., a cardiologist or a business analyst).

Noteworthy, A topic $t$ may not occur by its name/phrase explicitly in a document $d_i$. For example, a document about ``Mental Health'' may not include the exact phrase  ``Mental Health'', but still talk about ``Depression'', ``Anxiety'' and ``Antidepressant Drugs''. Thus, the topic ``Mental Health'' is implicit in this document, and it is equally important to annotate the implicit topics within the document and the explicit topics. Although the user-provided optional keywords may help mitigate this issue, it is almost impossible to provide a comprehensive list of keywords that can capture all possible ways ``Mental Health'' issues can be described. At the same time, a single appearance of a keyword may not always mean the document as a whole is focused on the corresponding topic. To summarize, neither the presence nor absence of keywords are sufficient to infer the correct topics associated with a document; they are just informative clues from the user end.

\section{\textbf{Method for Zero-shot Topic Inference}}\label{sec:Methods}
\vspace{-1mm}
\begin{figure*}
\centering
        \includegraphics[width=\linewidth,]{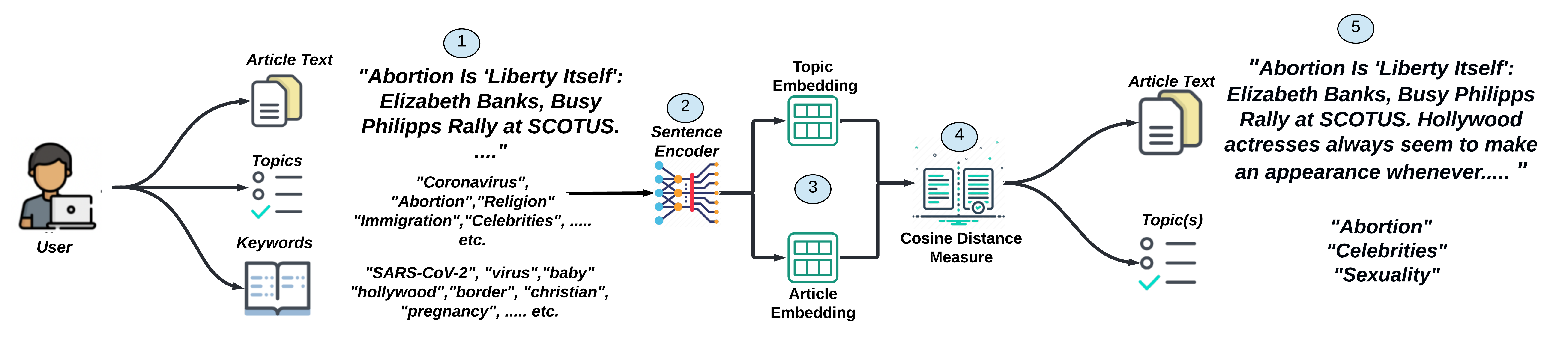}
        \vspace{-3mm}
        \caption{Steps for \textit{Zero-shot Multi-Label Topic Inference} process, leveraging sentence encoders.}
        \vspace{3mm}
        \label{fig:ZS-Topic Inference}
        \vspace{-3mm}
\end{figure*}

In this section, we discuss the \textit{zero-shot topic inference} approach we studied in this paper. The inputs are a corpus of documents, a set of user-defined topics, and an optional keyword list for each topic, whereas the output is each document labeled with zero to more topics. The end-to-end inference process is shown in Fig~\ref{fig:ZS-Topic Inference} and briefly described below.

\begin{enumerate}[leftmargin=*,itemsep=0ex,partopsep=-0.5ex,parsep=0ex]
\vspace{-1mm}
\item The end user provides the inputs, i.e., article text, custom-defined topics, and optional keywords.

\item Input article, topics, and keywords are fed to the sentence encoder model separately, and this is where we used different sentence encoders (InferSent, LASER, Universal Sentence Encoder (USE),  and Sentence-BERT).\label{sec:step2}

\item Next, Two separate embedding vectors are generated by sentence encoders:

\begin{itemize}[leftmargin=*,itemsep=0ex,partopsep=-0.5ex,parsep=0ex]

\item \underline{Article Embedding}: The input article is embedded by sentence encoders using three different approaches, the details of which are discussed in section \ref{sec:article_Embedding}.

\item \underline{Topic Embedding}: The candidate topics are embedded by combining the sentence embedding vectors of the topics as well as 
optional keywords/definitions/explicit mentions. In total, four different approaches have been explored; the details are provided in section~\ref{sec:topic_Embedding}.

\end{itemize}

\item Once we obtain these two embeddings, in the next step, we assess the semantic similarity between the two embeddings. Foremost, the semantic similarity is quantified using the cosine similarity between 2 embeddings. Then, based on the cosine similarity and a user-defined threshold, we assign topics to the article which are higher than the threshold. To do an exhaustive analysis, we experimented over a range of thresholds between 0 to 1. \label{sec:step4}

\item The output of the \textit{zero-shot topic inference} framework is the set of inferred topic(s). \label{sec:step5}
\end{enumerate}
 
 \vspace{-2mm}

\subsection{\textbf{Article Embedding}}\label{sec:article_Embedding}

\begin{table}[!htb]\small
        \centering
        \begin{tabular}{p{1.5cm}|p{5.5cm}}
        \hline
            {\textbf{Embedding Approach}} &  {\textbf{Description}}  \\
        \hline
        \hline
        Entire Article (EA) & 
            Encode the entire article using Sentence Encoders at once, including articles that are long paragraphs and consist of more than one sentence.\\
            \hline
        Sentence Embedding Average (SEA) &
             Split the article into sentences, then encode each sentence, and at the end, average all sentence embedding to generate article embedding.\\
            \hline
        Individual Sentence Embedding (ISE) & 
            Split the input article into sentences and encode each sentence separately. Then, unlike averaging (Sentence Embedding Average), use the individual sentence embeddings for similarity calculation with topic embedding.\\
            \hline

        \end{tabular}
        \caption {Three different ways of encoding an input article using sentence encoders.}
        \label{table:article_embed}
        
    \vspace{-4mm}
    \end{table}
    
For article embedding, we adopted three methods narrated in Table~\ref{table:article_embed}. As mentioned earlier, in our work, we leveraged contemporary sentence encoders for the mentioned task, such as:

\begin{enumerate}[leftmargin=*,itemsep=0ex,partopsep=-0.5ex,parsep=0ex]

\item InferSent~\cite{infersent}
\item Language-Agnostic SEntence Representations (LASER)~\cite{laser}
\item Sentence-BERT (SBERT)~\cite{sbert}
\item Universal Sentence Encoder (USE)~\cite{cer2018universal}
\end{enumerate}

We would like to mention that we did not perform fine-tuning or parameter tuning on top of the pre-trained sentence encoders. We have provided brief descriptions of the above encoders in appendix \ref{appen:encoders}.

\subsection{\textbf{Topic Embedding}}\label{sec:topic_Embedding}

For generating topic embedding, we adopted four approaches, including and excluding the auxiliary information provided by the user, to do a comparative study. The details of topic embedding are given in Table \ref{table:topic_embed}. As part of topic embedding using auxiliary information (Embedding approach ``Explicit-Mentions''), we performed a rudimentary annotation on the dataset to find explicit mentions of the topics, which is discussed in algorithm~\ref{alg:rudimentary}.

\begin{table}[!htb]\small
        \centering
        \begin{tabular}{m{1.7cm}|m{5cm}}
        \hline
            {\textbf{Embedding Approach}} &  {\textbf{Description}}  \\
        \hline
        \hline
        \textbf{Topic-Name-Only} & 
            Encode only the topic name/phrase. \\
            \hline
        \textbf{Topic + Keywords} &
             Encode both topic name and keywords, then average all embeddings to generate the final topic embedding. \\
            \hline
        \textbf{Topic + Keyword }\textbf{+ Definition} & 
            Extract the topic's and keyword's definitions from WordNet, encode these definitions separately using sentence encoders, and then average all embeddings to generate the final topic embedding. For example, instead of encoding the keyword ``campaign'', we generated embedding of its definition, ``a race between candidates for elective office''. \\
            \hline
       \textbf{Explicit-Mentions} & 
            First, extract all the articles explicitly mentioning the topic/phrase using algorithm \ref{alg:rudimentary} for all topics. Then, for each topic, generate embeddings of all articles which are explicitly annotated/labeled with that topic, then average them to obtain the ultimate topic embedding. \\
            \hline

        \end{tabular}
        \caption {Four different ways of encoding a topic using sentence encoders.}
        \label{table:topic_embed}
        \vspace{-4mm}
    \end{table}

\begin{algorithm}[!htb]\small
\caption{Article Annotation using Explicit Mention}
\label{alg:rudimentary}
\begin{algorithmic}[1]
\State \textbf{Input:} Article text, Topic names and Keywords
\State \textbf{Output:} Articles labeled with explicit topics
\For {each article text}
\State   check whether the topic name or set (at least 3) of the informative keywords are present or not in the corresponding article text 
\If {present}
   {label the article with the explicit topic\;}
\EndIf
\EndFor
\end{algorithmic}
\end{algorithm}

\subsection{\textbf{Zero-shot Topic Inference}}\label{sec:zero-shot}

Once we obtain all the embeddings, we measure cosine similarity between article and topic embeddings and, accordingly, infer topics. For instance, considering article $a$ and topics $t\in T$ as well as considering ``Entire Article'' (EA) and ``Topic-Name-Only'' (TNO) as the embedding approach for article and topic, respectively, the inference of topic works as follows:

\vspace{-4mm}
\[\hat{t} = {\mathrm{\underset{t \in T}{argmax} \left\{cosine\_similarity \left(EA(a), TNO(t) \right)\right\}}}\]
\vspace{-4mm}

Where topic $t$ belongs to a set of input topics $T$. $EA(a)$ represents the embedding of article $a$ using the ``Entire Article'' embedding approach, and $TNO(t)$ represents the embedding of topic $t$ using the ``Topic-Name-Only'' embedding approach. Note that, the combination of other articles and topic embeddings can be expressed in a similar way and hence, omitted due to lack of space.

\section {Experimental Design}\label{sec:Experiment}

\subsection{Datasets}\label{sec:data-sets}
Although the idea of our goal task is inspired from \textit{Definition-Wild Zero-Shot-Text Classification} coined by~\citet{yin2019benchmarking}, we realized that the dataset introduced in the paper is suitable for Single-Label Multi-class classification, whereas our zero-shot topic inference setup is a Multi-label classification problem (more than one topic is associated with the input article). Hence in our experiments, we mainly focused on curating/using the following datasets. (A) \textit{Large datasets} with a higher number of articles for inference and relatively longer text (News and Medical-Blog collected from the web), and (B) \textit{Small datasets} which contain fewer articles (<2000) for inference and are relatively shorter in length.

\textbf{\textit{Large Datasets}:} The large datasets were published by~\citet{sarkar2022concept},
which are collection of publicly available online news\footnote{https://newsbusters.org/} and medical-blog articles\footnote{https://www.health.harvard.edu/}. Each article is already labeled with one or more ground-truth topics and stored in JSON objects. Some statistics about these datasets are summarised in Table~\ref{table:Dataset}.

\textbf{\textit{Small Datasets}:} The small datasets are originally a set of 5 different online product reviews; these were initially collected from ~\citet{hu2004mining} and re-annotated by ~\citet{karmaker2016generative}. Unlike large datasets, the product reviews are shorter in length and contain more topics than the larger two datasets (see Table~\ref{table:Dataset}).

\begin{table}[!htb]\Large
\begin{adjustbox}{width=\columnwidth,center}
\centering
\begin{tabular}{ c c c c c}
\hline 
 {\textbf{Dataset}}  & {\textbf{Articles}} & {\textbf{Avg. Article}}  & {\textbf{Topics}}  & {\textbf{Topics/}}\\
   &  & {\textbf{length}}  &  & \textbf{article}\\
\hline
Medical & 2066 & 693 &18 & 1.128  \\
News  & 8940 & 589 & 12 & 0.805   \\
Cellular phone & 587 & 16 &  23 & 1.058\\
Digital camera 1  & 642 & 18 & 24  & 1.069\\
Digital camera 2  & 380 & 17 & 20 & 1.039 \\
DVD player  & 839 & 15 & 23 & 0.781 \\
Mp3 player & 1811 & 17 & 21 & 0.956 \\

\hline
\end{tabular}
\end{adjustbox}
\caption{\label{table:Dataset} Statistics on \textit{Large} and \textit{Small} datasets }
\vspace{-2mm}
\end{table}

In zero-shot learning, the auxiliary information about topics is provided by the end user (e.g., domain experts) conducting the inference task in the form of keywords/textual descriptions. In this section, we have shown some topics and corresponding keywords details from the Medical dataset (Table~\ref{table:med_keyword}); due to lack of space, we provided more examples in the appendix \ref{appen:aux_info}.

\begin{table}[!htb]\small
        \centering
        \begin{tabular}{r|m{5cm}}
        \hline
            {\textbf{Topic Name}} &  {\textbf{Keywords}}  \\
        \hline
        \hline
        Addiction & 
            Opioids,
            Alcohol,
            Drug \\
            \hline
        Headache &
            Migraine,
            Sinus,
            Chronic pain \\
            \hline
        Heart Health & 
            Hypertension, 
            Stroke,
            Cardiovascular\\
            \hline
        Mental Health & 
            Depression, 
            Anxiety,
            Antidepressant \\
            \hline
        Women's Health &
            Pregnancy,
            Breast,
            Birth\\
            \hline

        \end{tabular}
        \caption {Topics and optional keywords from the Medical dataset }
        \label{table:med_keyword}
        \vspace{-6mm}
    \end{table}

\subsection{\textbf{Baseline Approaches}}
As baselines, we used constrained topic modeling and a classical word embedding-based inference approach. 

 \textbf{Generative Feature Language Models} (GFLM) were proposed by~\cite{karmaker2016generative}). The paper suggested an approach based on generative feature language models that can mine the implicit topics effectively through unsupervised statistical learning. The parameters are optimized automatically using an Expectation-Maximization algorithm. Details on the method have been discussed in the appendix \ref{appen:GFLM}. 
 
\textbf{ Classical word embeddings} are a popular way to encode text data into a dense real-valued vector representation. In order to implement a zero-shot classifier, we encoded both the input document and the target topics using pre-trained word embeddings and then, computed vector similarity between the input document encoding and each target topic encoding separately. The implementation of the classical word-embedding-based zero-shot approach is very similar to our setup (discussed in \ref{sec:Methods}); the differences are:

\begin{enumerate}[leftmargin=*,itemsep=0ex,partopsep=-0.5ex,parsep=0ex]

\item Instead of sentence encoders in step \ref{sec:step2}, pre-trained Glove embedding is used. 

\item Articles are represented in two different ways. \\
a) \textit{Average Sentence Level Embedding:} For each input article, we encode the article by averaging the pre-trained embeddings (e.g., Glove) of each word present in that article. \\
b) \textit{Dictionary of Word Embeddings}: Extract word embedding of all words in an article, and instead of taking the average, we save them individually as a key-value pair.
\item For semantic similarity assessment between Article and Topic embeddings, we used two different metrics: 1) Euclidean distance and 2) Cosine Similarity. 
\end{enumerate}
Rest of the process, i.e. step \ref{sec:step4} and step \ref{sec:step5} are the same as discussed in section \ref{sec:Methods}.

\subsection{\textbf{Evaluation Metric}}
To measure the performance of each \textit{zero-shot topic inference} approach, we use three popular metrics available in the literature: Precision, Recall, and $F_1$ score. First, for each article, the model inferred topic(s) were compared against the list of ``gold'' topic(s) to compute the true positive, false positive, and false negative statistics for that article. Then, all such statistics for all the articles in a dataset were aggregated and used to compute the final Precision, Recall, and micro-averaged $F_1$ score.

\section {Performance Analysis and Findings}\label{sec:performance}
In this section, we present the performance details of each sentence encoder over various types of article and topic encoding techniques (mentioned in Table~\ref{table:article_embed} and~\ref{table:topic_embed}), respectively. As part of the evaluation, we report the $F_1$ score for all sentence encoders and omit the Precision and Recall scores due to lack of space. Table \ref{table:baseline} contains the baseline results for \textit{Small} and \textit{Large} datasets. Table \ref{Performancetable1} shows the performance of \textit{Small datasets} for four types of topic embedding techniques and four sentence encoders. Note that articles in \textit{Small datasets} are mostly single sentences; hence, we only considered ``Entire Article'' as the article embedding for the \textit{Small datasets}. In contrast, Table \ref{Performancetable_new} provides details on \textit{Large datasets} for twelve combinations, including four types of topic embedding techniques and three types of article embeddings.

\begin{table}[!htb]\Large
\begin{adjustbox}{width={\linewidth}}
\centering
\begin{tabular}{r|cc|cccc}
\hline 
 &  & & \multicolumn{4}{c}{\textbf{Classical Embedding (Glove)}}\\\cline{4-7}
 {\textbf{Dataset}}  & {\textbf{GFLM}}  & {\textbf{GFLM}} & \textbf{Euclid.}  & \textbf{Cosine} & \textbf{Euclid.}  & \textbf{Cosine} \\
  &  \textbf{-S }&  \textbf{-W} & \textbf{Word} & \textbf{Word} & \textbf{Sent.} &  \textbf{Sent.} \\
\hline
\hline
\textbf{Medical} &  0.532 & 0.530 & 0.212 & 0.154 & 0.105 & 0.142\\
\textbf{News}  & 0.494 & 0.492 & 0.141 &	0.171 & 0.113 & 0.115\\
\textbf{Cellular phone }& 0.497 & 0.504 & 0.082 & 0.074 & 0.084 & 0.068\\
\textbf{Digital cam. 1}  & 0.460 & 0.471 & 0.120 &	0.118 &	0.142 & 0.127\\
\textbf{Digital cam. 2}  &  0.494 & 0.497 & 0.084 & 0.091 & 0.078 & 0.095\\
\textbf{DVD player}  & 0.473 & 0.486 & 0.096 &	0.100 &	0.096 &	0.108\\
\textbf{Mp3 player }& 0.509 & 0.514 & 0.058 &	0.066 &	0.053 &	0.069\\

\hline
\end{tabular}
\end{adjustbox}
\caption{\label{table:baseline} $F_1$ score for Topic Modeling based baselines, GFLM-S, GFLM-W and Classical Embedding based baselines, Euclidean Word, Euclidean Sentence, Cosine Word, Cosine Sentence.}

\vspace{2mm}

\begin{adjustbox}{width={\linewidth}}
\begin{tabular}{p{2cm}|r|ccccc}
\hline
\textbf{Topic} &  & \multicolumn{5}{c}{\textbf{Small Datasets}}\\\cline{3-7}

\textbf{Embed} &  \textbf{Sentence} & \textbf{Cellular} & \textbf{Digital} & \textbf{Digital} & \textbf{DVD} & \textbf{Mp3}\\
\textbf{-ding} & \textbf{Encoder} & \textbf{phone} &  \textbf{cam. 1} & \textbf{cam. 2 } &  \textbf{player} &  \textbf{player}\\

\hline
\hline
\multirow{4}{2cm}{\textbf{Topic-Name-Only}}
& InferSent & 0.079 & 0.065 &  0.077 & 0.046  & 0.065\\
& LASER & 0.091 & 0.087 & 0.101 & 0.076 & 0.097 \\
& SBERT & 0.418 & 0.427 & 0.520 & 0.295 & 0.373\\
& USE & \textbf{0.435} & \textbf{0.432} &  \textbf{0.579} & \textbf{0.379} & \textbf{0.424} \\

\hline
\multirow{4}{2cm}{\textbf{Topic + Keywords}}
& InferSent & 0.077 & 0.063 & 0.080 & 0.045 & 0.055\\
& LASER & 0.093  & 0.091 & 0.107 & 0.095 & 0.094 \\
& SBERT & \textbf{0.549} & \textbf{0.503} & \textbf{0.554} & \textbf{0.478} &  \textbf{0.433}\\
& USE & 0.511 & 0.477 & 0.501 & 0.442 & 0.398\\

\hline
\multirow{4}{2cm}{\textbf{Topic + Keyword + Definition}}
& InferSent & 0.091 & 0.086 & 0.083 & 0.061 &  0.091\\
& LASER & 0.192  & 0.212 &  0.100 & 0.247  & 0.165 \\
& SBERT & 0.220 & \textbf{0.273} & \textbf{0.321} & \textbf{0.325} & 0.277\\ 
& USE & \textbf{0.228} &  0.266 &  0.236 & 0.261 & \textbf{0.294}\\

\hline
\multirow{4}{2cm}{\textbf{Explicit-Mentions}}
& InferSent & 0.346 & 0.312 & 0.356 &  0.354 & 0.254\\
& LASER & 0.293 & 0.337 & 0.370 & 0.323 & 0.280\\
& SBERT &  \textbf{0.520} & \textbf{0.500} & \textbf{0.603} &  \textbf{0.501} & \textbf{0.521}\\
& USE & 0.488 & 0.457 & 0.593 & 0.449 & 0.486\\

\hline
\end{tabular}%
\end{adjustbox}

\caption{\label{Performancetable1}  $F_1$ score for the \textit{zero-shot topic inference} task for \textit{Small dataset}s (Cellular phone, Digital camera 1, Digital camera 2, DVD player, Mp3 player ). Performance comparison of four sentence encoders over various topic embedding procedures for "Article Embedding" type.}
\end{table}

\begin{table*}
\centering
\begin{adjustbox}{width=\linewidth,center}
\begin{tabular}{c | r| c c c | c c c | c c c |c c c }
\hline 
\multicolumn{2}{c|}{\textbf{Dataset ->}} & \multicolumn{12}{c}{\textbf{Medical }}\\
\hline
\multicolumn{2}{c|}{\textbf{Topic Embedding ->}} 
&\multicolumn{3}{c|}{\textbf{Topic Name Only}} 
& \multicolumn{3}{c|}{\textbf{Topic+Keywords}} 
& \multicolumn{3}{c|}{\textbf{Topic+Keyword+Def'n }} 
& \multicolumn{3}{c}{\textbf{Explicit-Mentions}} \\
\hline

\multicolumn{2}{c|}{\textbf{Article Embedding ->}}
& \textbf{EA} & \textbf{SEA} & \textbf{ISE} 
& \textbf{EA} & \textbf{SEA} & \textbf{ISE}  
& \textbf{EA} & \textbf{SEA} & \textbf{ISE}  
& \textbf{EA} & \textbf{SEA} & \textbf{ISE}  \\ \hline

\multirow{4}{1.7cm}{\textbf{Sentence Encoder}}
& InferSent & 0.128 & 0.146  & 0.120 & 0.102 & 0.105 & 0.119 & 0.140 & 0.132 & 0.131 & 0.154 & 0.217 & 0.227\\\
&  LASER & 0.120 & 0.142 & 0.134 & 0.124 & 0.122 & 0.121 & 0.125 & 0.124 & 0.185 & 0.187 & 0.139 & 0.136\\
&  SBERT & \textbf{0.565} & \textbf{0.571}  &\textbf{ 0.547} & \textbf{0.579}  & \textbf{0.541} & \textbf{0.471} & \textbf{0.460} & \textbf{0.465} & \textbf{0.420} & \textbf{0.594} & \textbf{0.556} & \textbf{0.534}\\
&  USE & 0.488 & 0.516 & 0.429 & 0.500 & 0.484 & 0.340 & 0.390 & 0.409 & 0.375 & 0.520 & 0.504 & 0.468\\
\hline

\multicolumn{2}{c|}{\textbf{Dataset ->}} & \multicolumn{12}{c}{\textbf{News }}\\
\hline
\multicolumn{2}{c|}{\textbf{Topic Embedding ->}} 
&\multicolumn{3}{c|}{\textbf{Topic Name Only}} 
& \multicolumn{3}{c|}{\textbf{Topic+Keywords}} 
& \multicolumn{3}{c|}{\textbf{Topic+Keyword+Def'n}} 
& \multicolumn{3}{c}{\textbf{Explicit-Mentions}} \\
\hline

\multicolumn{2}{c|}{\textbf{Article Embedding ->}}
& \textbf{EA} & \textbf{SEA} & \textbf{ISE} 
& \textbf{EA} & \textbf{SEA} & \textbf{ISE}  
& \textbf{EA} & \textbf{SEA} & \textbf{ISE}  
& \textbf{EA} & \textbf{SEA} & \textbf{ISE}  \\ 
\hline
\multirow{4}{1.7cm}{\textbf{Sentence Encoder}}
& InferSent & 0.105 & 0.116 & 0.099 & 0.217 & 0.127 & 0.110 & 0.129 & 0.141 & 0.117 & 0.234 & 0.161 & 0.144\\
&  LASER & 0.171 & 0.180 & 0.154 & 0.181 & 0.176 & 0.135 & 0.126 & 0.127 & 0.128 & 0.130 & 0.136 & 0.134\\
&  SBERT & \textbf{0.425} & 0.408 & \textbf{0.447} & \textbf{0.488} & \textbf{0.458} & \textbf{0.374} & 0.406 & 0.386 & 0.378 & \textbf{0.511} & \textbf{0.416} & \textbf{0.404}\\
&  USE & 0.419 & \textbf{0.426} & 0.367 & 0.461 & 0.418 & 0.281 & \textbf{0.420} & \textbf{0.390} & \textbf{0.391} & 0.446 & 0.371 & 0.368\\
\hline

\hline
\end{tabular}%
\end{adjustbox}

\caption{\label{Performancetable_new} F1-Score for the \textit{zero-shot topic inference} task for \textit{Large datasets} (Medical and News). Performance comparison of four sentence encoders over various topic embedding procedures and article embedding procedures.}

\end{table*}

 Below we have summarized our findings by analyzing Table~\ref{table:baseline},~\ref{Performancetable1} and ~\ref{Performancetable_new}:

\begin{enumerate}[leftmargin=*,itemsep=0.5ex,partopsep=0.0ex,parsep=0ex]
\item Overall, Sentence-BERT (SBERT) outruns all the encoders and baselines, GFLM-S, GFLM-W, and classical embedding based methods, for all datasets. USE performed close to SBERT; however, InferSent and LASER performed poorly over both datasets. For qualitative analysis of the classified data, we picked a review from the Digital camera 2 (\textit{Small}) dataset, which is associated with ground truth \textit{\textbf{"Size"}}, \textit{\textbf{"Lens"}}, \textit{\textbf{"Photo"}}. We observed that InferSent and LASER annotated the review with many incorrect topics, e.g. \textit{\textbf{"Design"}}, \textit{\textbf{"Feature"}}, \textit{\textbf{"Manual"}}, \textit{\textbf{"Weight"}}, \textit{\textbf{"Focus"}} etc. Universal Sentence Encoder (USE) annotated the same review with correct and some other topics which are semantically correlated to the correct topics, for instances \textit{\textbf{"Size"}} (highly correlated with \textit{\textbf{"Weight"}}), \textit{\textbf{"Focus"}}(highly correlated with \textit{\textbf{"Lens"}}), \textit{\textbf{"Video"}}(highly correlated with \textit{\textbf{"Photo"}}). On the other hand, for the same review, SentenceBERT inferred correct topics \textit{\textbf{"Size"}}, \textit{\textbf{"Lens"}}, \textit{\textbf{"Photo"}} and an incorrect topic \textit{\textbf{"Video"}}, thus achieve best $F_1$ Score among all the encoders.
Due to space limitation, we have added case study from \textit{Large} dataset in the appendix \ref{appen:case study}.
\item Even though USE could not beat SBERT generally, USE attained a score very close to the baseline methods (GFLM) and SBERT.

\item From the perspective of different topic embedding techniques, we observed that ``Topic+Keywords'' and ``Explicit-Mentions'' seemed to surpass the other two topic embedding techniques. Both of them include auxiliary information from end users  indicating that using user guidance (in the form of topic keywords) helps zero-shot topic inference in real time.

\item From the perspective of different article embedding techniques, mostly ``Entire Article'' appeared superior to others, with ``Sentence Embedding Average'' being the second best. Whereas, ``Individual Sentence Embedding'' was not promising.

\item ``Explicit-Mentions'' topic embedding with ``Entire Article'' as the article embedding attained the best score, followed by ``Topic + Keywords'' topic embedding paired with ``Entire Article''.

\item $F_1$ score obtained by InferSent and LASER indicates that these encoders failed to generalize over unseen datasets and, therefore, may not be a good choice for \textit{zero-shot topic inference}. 

\item Despite the observation stated in (6), we would like to point out that the inclusion of user guidance in the inference process boosted the performance of both InferSent and LASER. For example, ``Topic-Name Only'' embedding achieved around 7\% $F_1$ score (Average over all datasets). However, with "Explicit-Mentions" embedding, $F_1$ score elevated to around 30\% (Average over all datasets). 

\item For small datasets, ``Topic-Name-Only'' embedding presented an interesting case. Here, USE performed better than SBERT. This suggests that, for the product review domain, if additional keywords for each topic are unavailable, USE may be a better choice than SBERT. However, a detailed investigation is warranted to determine the root cause for this result.

\end{enumerate}
\vspace{-1mm}

\begin{table*}\Large
\centering
\begin{adjustbox}{width=\linewidth}
\begin{tabular}{r|cccc|cccc|cccc|cccc}

\hline
&\multicolumn{4}{c|}{\textbf{Topic Name Only}} 
& \multicolumn{4}{c|}{\textbf{Topic+Keywords}} 
& \multicolumn{4}{c|}{\textbf{Topic + Keyword + Definition}} 
& \multicolumn{4}{c}{\textbf{Explicit-Mentions}} \\
\hline

{\textbf{Encoder}}
& \textbf{Infer.} & \textbf{LASER} & \textbf{SBERT} & \textbf{USE}
& \textbf{Infer.} & \textbf{LASER} & \textbf{SBERT} & \textbf{USE}
& \textbf{Infer.} & \textbf{LASER} & \textbf{SBERT} & \textbf{USE}
& \textbf{Infer.} & \textbf{LASER} & \textbf{SBERT} & \textbf{USE} \\ 
\hline

\textbf{Medical} & 0.720  & 0.602 & 0.117 & 0.704 & 1.964 & 1.024 & 1.720 & 2.183 & 7.963 & 2.353 & 1.822 & 2.082 & 730.655 & 708.560 & 19.453 & 21.154\\
\textbf{News} & 0.391 & 0.797 & 0.080 & 0.760 & 1.673 & 0.801 & 1.479 & 0.809 & 1.786 & 1.066 & 1.455 & 1.215 & 2885.003 & 1438.609 & 61.226 & 36.005\\
\textbf{Cellular phone} & 0.350 & 0.147 & 0.118 & 0.706 & 0.622 & 0.456 & 0.587 & 0.985 & 2.707 & 0.790  & 0.570 & 1.043 & 10.323 & 6.726 & 7.540 & 9.025 \\
\textbf{Digital cam. 1} & 0.302 & 0.408 & 0.132 & 0.802 & 0.621 & 0.490 & 0.416 &  0.914 & 1.913 & 0.689 & 0.384 & 0.948 & 13.002 & 9.707 & 9.597 & 8.865\\
\textbf{Digital cam. 2 } & 0.418 & 0.073 & 0.109 & 0.721 & 0.659  & 0.405 & 0.313 & 0.846 & 1.354 & 0.534 & 0.535 & 1.159 & 6.780 & 5.844 & 4.665 & 5.669 \\
\textbf{DVD player} & 0.450 & 0.360 & 0.114 & 0.795 &  0.682 & 0.422  & 0.373 & 0.833 & 1.673 & 0.852 & 0.385 & 0.816 & 9.695 & 10.822 & 6.802 & 7.271\\
\textbf{Mp3 player} & 0.466 & 0.190 & 0.122 & 0.762 & 0.574 & 0.707 & 0.775 & 1.167  & 2.654 & 0.963 & 0.729 & 1.217 & 29.245 & 18.380 & 19.833 & 21.239\\

\hline
\end{tabular}%
\end{adjustbox}

\caption{\label{Performancetable_time1} Time comparison for generating topic embedding by different sentence encoders (Time unit in seconds).}
\vspace{-2mm}

\end{table*}

\begin{table}
\centering

\begin{adjustbox}{width={\linewidth}}
\begin{tabular}{r|ccccc}

\hline
\multicolumn{6}{c}{\textbf{Total Time for Computing Embedding for Entire Article}}\\\hline
\hline
& \multicolumn{5}{c}{\textbf{Small Datasets}}\\\cline{2-6}
\textbf{Sentence} & \textbf{Cellular} & \textbf{Digital} & \textbf{Digital} & \textbf{DVD} & \textbf{Mp3}\\
\textbf{Encoder} & \textbf{phone} &  \textbf{cam. 1} & \textbf{cam. 2 } &  \textbf{player} &  \textbf{player}\\
\hline

InferSent & 8.645 & 7.819 & 4.610 & 10.668 & 21.433\\
LASER & 6.952 & 8.267 & 4.451 & 9.989  & 15.387 \\
SBERT & 6.790 & 6.973 & 4.454 &  9.634 & 18.783\\
USE & 5.618 & 6.739 & 4.091 & 8.292 & 17.496 \\

\hline
\end{tabular}%
\end{adjustbox}
\label{Performancetable3}
\vspace{-3mm}
\end{table}

\begin{table}[!htb]
\centering
\begin{adjustbox}{width={0.8\linewidth}}
\begin{tabular}{p{2cm} | r | r | r  }
\hline
\multicolumn{4}{c}{\textbf{Total Time for Computing Article Embedding}}\\\hline\hline

\textbf{Article} & \textbf{Sentence} & \multicolumn{2}{c}{\textbf{Large Datasets}}\\\cline{3-4}
\textbf{Embedding} & \textbf{Encoder} & \textbf{Medical} & \textbf{News} \\

\hline
\multirow{4}{2cm}{\textbf{Entire Article}}
& InferSent &  902.851 &	3867.350\\
& LASER &   514.929	& 1919.147\\
& SBERT &   28.805 & 88.633\\
& USE &   27.478 &	64.203\\

\hline
\multirow{4}{2cm}{\textbf{Sentence Embedding Average}}
& InferSent &  1035.469 &	3807.204\\
& LASER &   639.066 &	2373.594	\\
& SBERT &   548.573	& 1891.539\\
& USE &   412.942 &	1448.037\\

\hline
\multirow{4}{2cm}{\textbf{Individual Sentence Embedding}}
& InferSent &  1022.728	 & 3778.628\\
& LASER &   631.533	 & 2350.273\\
& SBERT &   553.106	& 1876.776\\
& USE &   428.725 &	1410.522\\

\hline

\end{tabular}%
\end{adjustbox}

\caption{\label{PerformanceTable_time2} Time comparison for generating article embedding by different sentence encoders for \textit{Small and Large datasets. (Time unit in seconds)}  }
\vspace{-3mm}
\end{table}

For analyzing computation time, we also logged the time taken by different encoders for different types of article and topic embeddings. Since our goal task focuses on the real-time scenario, an important criterion for picking the right approach is the inference time. Therefore, we have reported embedding generation time (in seconds) for each model in Tables \ref{Performancetable_time1} and \ref{PerformanceTable_time2}. Some observations from these tables are as below.

\begin{enumerate}[leftmargin=*,itemsep=0.5ex,partopsep=0.0ex,parsep=0ex]

\item USE is the fastest of all encoders for generating sentence embedding, followed by SBERT. 

\item Among all topic embedding techniques, ``Explicit Mentions'' took the highest time for processing since, for ``Explicit Mentions'', the encoder needs to traverse the whole dataset.

\item Among all article embedding techniques, ``Entire Article'' 
appeared to be the fastest for USE and SBERT. In particular, for ``Entire Article'' embedding on the News dataset USE took approximately 64 seconds, whereas, InferSent and LASER took around 3867 and 1919 seconds, respectively.

\item The difference in article embedding time is more conspicuous on the \textit{Large datasets} as they contain a longer and higher number of articles. USE, SBERT clearly wins over InferSent and LASER in comparison of time as well.

\item The high processing time over \textit{Large datasets} suggests that InferSent and LASER are unsuitable for real-time inference if the dataset is vast and comprised of long articles.
\end{enumerate}

In essence, comprehensive performance and run-time analysis show that a) auxiliary information helps in achieving better performance in real-time \textit{zero-shot topic inference} task, b) even though the sentence encoders are designed to be fairly general, aiming for seamless transfer learning, not all of them serve the purpose accurately, c) the processing time varies a lot across different sentence encoders and should be considered seriously while using these encoders in real-time tasks.

\section{Conclusion}\label{section:conclusion}

\textit{Zero-shot topic inference} is a fundamental yet challenging task. Topic inference task tries to identify submerged topics from documents in a given corpus and is proven easier to solve in a supervised problem setup. However, the recent shift towards zero-shot and transfer learning due to expensive training data motivated us to inspect the task from a zero-shot perspective. Since \textit{zero-shot topic inference} is a much more difficult problem and recent sentence encoders are relatively under-explored in this context, we leveraged four popular sentence encoders to examine their generalization power for the task. 

Our \textit{zero-shot topic inference} task formulation demands serving end users in real time. Although popular sentence encoders have been shown to achieve better generalization over many downstream NLP tasks, not all performed well for the \textit{zero-shot topic inference} task. Amidst four encoders, Sentence-BERT seemed to perform well on unseen data, while USE performed decently and was the second best. However, InferSent and LASER did not achieve a comparable performance against USE and SBERT. We also proposed novel ways for incorporating user guidance in the zero-shot process, which improved the overall accuracy of topic inference. Apart from performance on unseen datasets, we compared execution time which revealed that not only the $F_1$ score but also the execution time poses a reservation on the use of InferSent and LASER for the mentioned task.

\bibliography{custom}

\appendix

\section{Appendix}
\label{sec:appendix}

\subsection{Limitations}\label{appen:limitation}
Despite the promising results, we feel a limitation of our work is the reliance on keywords, i.e., the performance of the real-time zero-shot greatly depends on the choice of keywords; without appropriate keywords, the approach may suffer. In our future work, we will work towards mitigating this constraint. We believe that ever-increasing scale of the data in different areas, new types of contents, topics, etc. will encourage the community to focus more towards \textit{zero-shot topic inference} for categorization, annotation and also motivate researchers to pursue research in this important direction.  

\subsection{Auxiliary Information Generation}\label{appen:aux_info}
Keyword list for each topic (based on the user's experience and expectations) was not readily available, for which we ideally needed a real user. To address this limitation, we followed the steps discussed in ~\citet{sarkar2022concept}. We extracted the \textit{informative} keywords for each topic using the TF-IDF heuristics. Then, a set of keywords for each topic were selected through careful inspection of the top keywords with high TF-IDF scores.
In Tables \ref{table:news_keyword} and \ref{table:creative_keyword}, we have shown some topics and corresponding keywords details from News and Mp3 player dataset respectively.


\begin{table}[!htb]\small
        \centering
        \begin{tabular}{r|m{4.5cm}}
        \hline
        {\textbf{Topic Name}} &  {\textbf{Keywords}}\\
        \hline
        \hline
        Economy &
            Recession,
            Budget,
            Stock Market\\
            \hline
        Global Warming & 
            Climate, 
            Planet, 
            Green \\
            \hline
        Immigration & 
            Border,
            Immigrants,
            Detention\\
            \hline
        Religion &
            Christian,
            Religious,
            Church\\
            \hline
        Sexuality & 
            Gay,
            Lgbtq,
            Transgender\\
            \hline
        \end{tabular}
        \caption { Topics and optional keywords from News dataset }
        \label{table:news_keyword}
        
        \centering
        \begin{tabular}{r|m{4.5cm}}
        \hline
        {\textbf{Topic Name}} &  {\textbf{Keywords}}\\
        \hline
        \hline
        Screen &
            Display,
            Screen saver,
            Interface\\
            \hline
        Sound & 
            Audio, 
            Headphone, 
            Earbud \\
            \hline
        Navigation & 
            Control,
            Scroll,
            Flywheel\\
            \hline
        Battery &
            Power,
            Recharge,
            mAh\\
            \hline
        \end{tabular}
        \caption{Topics and optional keywords from Mp3 player dataset}
        \label{table:creative_keyword}
        \vspace{-6mm}
    \end{table}

\subsection{Details on Generative Feature Language Models}\label{appen:GFLM}

Generative Feature Language Models (GFLM) is unsupervised statistical learning in which parameters are optimized automatically using an Expectation-Maximization algorithm. Once the EM algorithm converges, one knows the topic distributions, i.e. 

\begin{enumerate}

\item $P(z_{D,w} =t)$: Contribution of topic $t$ for the generation of a particular word.
\item $P(z_{D,w}=B)$: Contribution of background model (mostly stop-words) for the generation of a particular word. 
\item $\pi_{D,t}$: to what proportion, a particular document $D$ is generated from some topic-of-interest $t$.

\end{enumerate}

Based on these quantities, topic distributions within various documents can be inferred in two different ways, which were called {\bf GFLM-Word} (GFLM-W) and {\bf GFLM-Sentence} (GFLM-S).


\textbf{GFLM-Word:} It looks at each word $w$ in the document $D$ and adds a topic $t$ to the inferred topic list if and only if $p(z_{D,w} =t) \times (1-p(z_{D,w}=B))$ is greater than some threshold $\theta$ for at least one word in $D$. The philosophy behind this formula is that if any particular word $w$ has a small probability of being generated by a background model but has a higher probability of being generated from some topic $t$, then word $w$ is likely referring to topic $t$. Here, the decision is made solely by looking at individual words, not the entire document.

\textbf{GFLM-Sentence:} Given a document $D$, it looks at the contribution of each topic $t$ in the generation of the sentence, i.e., $\pi_{D,t}$  and infers $t^*$ as the topic only if $\pi_{D,t^*}$ is greater than some user-defined threshold $\theta$. Here, the decision is made at the sentence level, not at the word level.

\subsection{Sentence Encoders}\label{appen:encoders}
This section presents a bird's-eye view of the sentence encoders we have used for our experiments.

\textbf{InferSent} ~\citet{infersent} was released by Researchers at Facebook, which employs a supervised method to learn sentence embeddings. It was trained on natural language inference data and generalizes well to many different tasks\footnote{To use InferSent encoder we download the state-of-the-art fastText embedding and also download the pre-trained model from https://dl.fbaipublicfiles.com/senteval/infersent/infersent2.pkl.}. They found that models learned on NLI tasks can perform better than models trained in unsupervised conditions or on other supervised tasks \citet{conneau2017supervised}. Furthermore, by exploring various architectures, they showed that a BiLSTM network with max-pooling outperformed the state-of-the-art sentence encoding methods, outperforming existing approaches like SkipThought vectors \citet{kiros2015skip}. The model encodes text in 4,096 dimensional vectors.

\textbf{ Language-Agnostic Sentence Representations (LASER)}~\citet{laser}, is a method to generate pre-trained language representation in multiple languages. It was released by Facebook. LASER architecture is the same as neural machine translation: an encoder/decoder approach. It uses one shared encoder for all input languages and a shared decoder to generate the output language. The encoder is a five-layer bidirectional LSTM network. It does not use an attention mechanism but, has a 1,024-dimension vector to represent the input sentence. It is obtained by max-pooling over the last states of the BiLSTM, enabling comparison of sentence representations.

\textbf{Sentence-BERT (SBERT)}~\citet{reimers2019sentence}, is a modification of the pre-trained BERT network that use siamese and triplet network structures to derive semantically meaningful sentence embeddings that can be compared using cosine-similarity. 
SBERT is a so-called twin network that allows it to process two sentences in the same way, simultaneously. BERT makes up the base of this model, to which a pooling layer has been appended. This pooling layer enables to create a fixed-size representation for input sentences of varying lengths. Since the purpose of creating these fixed-size sentence embeddings was to encode their semantics, the authors fine-tuned their network on Semantic Textual Similarity data. They combined the Stanford Natural Language Inference (SNLI) dataset with the Multi-Genre NLI (MG-NLI) dataset to create a collection of 1,000,000 sentence pairs. The training task posed by this dataset is to predict the label of each pair, which can be one of “contradiction”, “entailment” or “neutral”.

In 2018, Researchers at Google released a \textbf{Universal Sentence Encoder (USE)}~\citet{cer2018universal} model for sentence-level transfer learning that achieves consistent performance across multiple NLP tasks. The models take as input English strings and produce as output a fixed dimensional (512) embedding representation of the string. To learn the sentence embeddings, the encoder is shared and trained across a range of unsupervised tasks along with supervised training on the SNLI corpus for tasks like a) Modified Skip-thoughtPermalink, b) Conversational Input-Response PredictionPermalink, c) Natural Language Inference. there are two architectures proposed for USE (Transformer Encoder and Deep Averaging Network). Based on shorter inference time observed in our experiments, we used Deep Averaging Network (DAN)\footnote{
https://tfhub.dev/google/universal-sentence-encoder/4} architecture.

\subsection{Case Study}\label{appen:case study}

Due to space restriction we could not present case studies in the main paper. Hence we have added our observation in this sections:

 \textbf{Case study on \textit{Large Datasets}:} Upon qualitative analysis of the classified data from Medical dataset, we observed while the input was an article which is originally labeled with topics \textit{\textbf{"Heart Health"}} and \textit{\textbf{"Mental Health"}}; InferSent and LASER seemed to infer the correct topics along with many incorrect topics such as \textit{\textbf{"Brain and cognitive health"}}, \textit{\textbf{"Healthy Eating"}}, \textit{\textbf{"Women's Health"}}, \textit{\textbf{"Children's Health}"} etc. The annotated dataset clearly indicates that InferSent and LASER are unable to distinguish between the topics in a zero-shot setting. \\
Universal Sentence Encoder (USE) annotated the same article as \textit{\textbf{"Heart Health"}}, \textit{\textbf{"Mental Health"}} and \textit{\textbf{"Brain and cognitive health"}}. Correlation analysis reveals that \textit{\textbf{"Mental Health"}} and \textit{\textbf{"Brain and cognitive health"}} have high semantic correlation and therefore USE inferred both the topics. \\
Compared to all these sentence encoders, SentenceBERT performed precisely and inferred the correct topics for the article mentioned earlier. \\
The case study also corroborate our observations discussed in the paper that InferSent and LASER are unsuited for zero-shot approaches. While Universal Sentence Encoder performs moderate except for semantically correlated topics. SentenceBERT outrun all these encoders and effectively annotate datasets in the zero-shot approaches.

\end{document}